\documentclass[10pt,twocolumn,letterpaper]{article}

\usepackage{wacv}              

\usepackage{cuted}
\usepackage{float}
\usepackage{capt-of}

\usepackage{epsfig}
\usepackage{graphicx}
\usepackage{amsmath}
\usepackage{amssymb}
\usepackage{booktabs}
\usepackage{appendix}
\usepackage{arydshln}
\usepackage{soul}

\usepackage{import}
\usepackage{xcolor}         
\usepackage{color,soul}     
\usepackage{makecell}
\usepackage{multirow}
\usepackage{enumitem}
\usepackage{graphicx}
\usepackage[accsupp]{axessibility}  

\definecolor{blue_munsell}{rgb}{0.36, 0.54, 0.66}
\definecolor{blue-violet}{rgb}{0.54, 0.17, 0.89}
\definecolor{byzantine}{rgb}{0.74, 0.2, 0.64}
\definecolor{caputmortuum}{rgb}{0.35, 0.15, 0.13}
\definecolor{other}{RGB}{83, 190, 198}

\usepackage[pagebackref,breaklinks,colorlinks]{hyperref}

\newcommand\blfootnote[1]{%
  \begingroup
  \renewcommand\thefootnote{}\footnote{#1}%
  \addtocounter{footnote}{-1}%
  \endgroup
}

\newcommand*{\imagenet}{\textsc{ImageNet}\xspace}
\newcommand*{\resnet}{\textsc{ResNet}\xspace}
\newcommand*{\resnets}{\textsc{ResNet}s\xspace}

\newcommand*{\resnetOneFiftyTwo}{\textsc{ResNet152}\xspace}

\newcommand*{\deit}{\textsc{DeiT}\xspace}

\newcommand*{\deitbase}{\textsc{DeiT-B}\xspace}

\newcommand*{\aptos}{\textsc{APTOS2019}\xspace}
\newcommand*{\ddsm}{\textsc{DDSM}\xspace}
\newcommand*{\isic}{\textsc{ISIC}\xspace}
\newcommand*{\chexpert}{\textsc{CheXpert}\xspace}

\newcommand*{\dino}{\textsc{Dino}v1\xspace}
\newcommand*{\dinotwo}{\textsc{Dino}v2\xspace}
\newcommand*{\dinos}{\textsc{Dino}\xspace}
\newcommand*{\sam}{\textsc{Sam}\xspace}
\newcommand*{\seem}{\textsc{Seem}\xspace}
\newcommand*{\clip}{\textsc{OpenCLIP}\xspace}
\newcommand*{\ImNet}{\textsc{ImNet-1k-Sup}\xspace}
\newcommand*{\vit}{\textsc{ViT}\xspace}
\newcommand*{\orgclip}{\textsc{CLIP}\xspace}
\newcommand*{\coco}{\textsc{COCO2017}\xspace}
\newcommand*{\refcoco}{\textsc{Ref-COCO}\xspace}
\newcommand*{\laion}{\textsc{laion-5B}\xspace}
\newcommand*{\laiontwob}{\textsc{laion-2B-EN}\xspace}
\newcommand*{\laionfour}{\textsc{laion-400M}\xspace}
\newcommand*{\blip}{\textsc{BLIP}\xspace}
\newcommand*{\cnn}{\textsc{CNN}\xspace}

\def\wacvPaperID{0287} 
\def\confName{WACV}
\def\confYear{2024}

\begin{document}

\title{Are Natural Domain Foundation Models Useful for Medical Image Classification?}

\author{Joana Palés Huix \textsuperscript{
$1,2,3$}
\hspace{-1.5mm}
\thanks{Corresponding author: Joana Palés Huix \textless{}joanaph@kth.se\textgreater{}} 
\hspace{0.05mm}
\thanks{These authors contributed equally to this work. \vspace{-1.75mm}}
\qquad
Adithya Raju Ganeshan \textsuperscript{$1,2,\dagger$}
\qquad
Johan Fredin Haslum \textsuperscript{$1,2,3$} \\
\qquad
Magnus Söderberg \textsuperscript{$3$}
\qquad
Christos Matsoukas \textsuperscript{$1,2,3$}
\qquad
Kevin Smith \textsuperscript{$1,2$}
\\\\
\textsuperscript{$1$} KTH Royal Institute of Technology, Stockholm, Sweden 
\textsuperscript{$2$} Science for Life Laboratory, \\ Stockholm, Sweden 
\textsuperscript{$3$}  AstraZeneca, Gothenburg, Sweden 
}
\maketitle

\blfootnote{ \textit{ \hrule \vspace{1mm} \hspace{-3.75mm}
Originally published at the Winter Conference on Applications of Computer Vision (WACV 2024).}}

\begin{abstract}
The deep learning field is converging towards the use of general foundation models that can be easily adapted for diverse tasks. While this paradigm shift has become common practice within the field of natural language processing, progress has been slower in computer vision. In this paper we attempt to address this issue by investigating the transferability of various state-of-the-art foundation models to medical image classification tasks. Specifically, we evaluate the performance of five foundation models, namely \sam, \seem, \dinotwo, \blip , and \clip across four well-established medical imaging datasets. We explore different training settings to fully harness the potential of these models.
Our study shows mixed results. 
\dinotwo consistently outperforms the standard practice of \imagenet pretraining. However, other foundation models failed to consistently beat this established baseline indicating limitations in their transferability to medical image classification tasks.

\end{abstract}
\section{Introduction}
\label{sec:intro}

Recently there has been a surge of interest in the use of foundation models trained on large-scale datasets within the field of computer vision. 
This has resulted in a growing trend of adapting these models for a wide range of downstream applications with minimal effort. 
\begin{figure}[!ht]
\begin{center}
   \includegraphics[width=\linewidth]{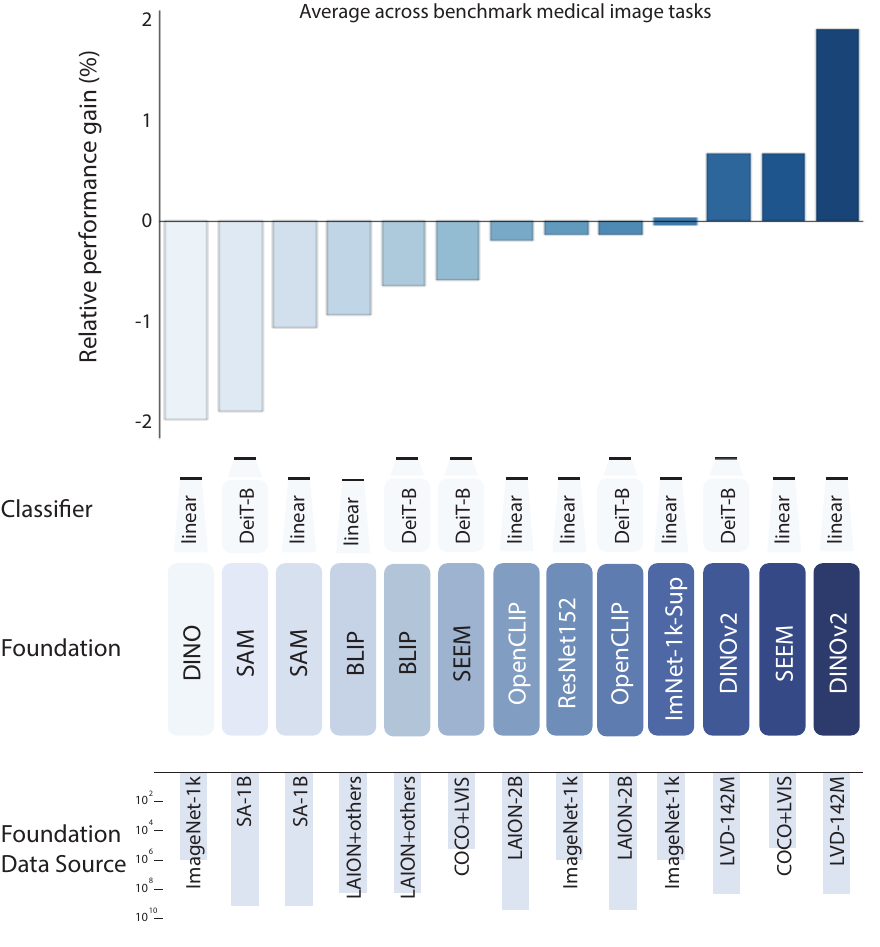}
\end{center}
\vspace{-5mm}
\caption{Foundation models applied to benchmark medical image classification tasks. We report different combinations of foundation models (linear heads or Vision Transformers (\vit) with linear heads) and classifiers for relative gain in performance for each setting. We also provide the size of the source data on which the foundation models were trained, ranging from LVD+COCO ($2\times10^5$) to LAION-2B ($2\times10^9$). The baseline consists of a \vit foundation pretrained on \imagenet-1k in a supervised manner and fine-tuned for the medical task. While some combinations perform below the baseline, \dinotwo consistently outperforms standard \imagenet pretraining.}
\label{fig:Fig1}
\vspace{-6mm}
\end{figure}

Foundation models, characterized by their substantial size and self-supervised training on diverse datasets, possess remarkable capabilities for generating meaningful representations across multiple domains \cite{bommasani_opportunities_2022}. These models offer significant advantages as parameter initialization strategies for a wide range of downstream tasks.
While the concept of large-scale pretraining originated in the field of Natural Language Processing (NLP) with BERT \cite{devlin2018bert}, subsequent advancements like BART \cite{lewis2019bart}, RoBERTa \cite{liu2019roberta}, GPT \cite{radford2018improving,radford2019language,brown_language_2020}, have further propelled the utilization of task-agnostic models. 
This paradigm shift has not only driven research on adaptation methods but has also fueled a growing interest in comprehending the inner workings of foundation models \cite{zhou2023comprehensive, bommasani_opportunities_2022}.

The field of computer vision is arguably currently undergoing a similar transformation. 
Until recently, computer vision relied heavily on models pretrained on \imagenet \cite{deng2009imagenet} using supervised learning. 
However, recent advancements have led to the emergence of alternative computer vision foundation models using larger datasets and self-supervision in a move that diverges from \imagenet as the primary source of pretraining data. 
These models can be broadly categorized into two types: feature encoding models trained on pretext tasks, such as \dinos \cite{oquab_dinov2_2023, caron2021emerging}, \orgclip \cite{radford2021clip} and \blip \cite{li2022blip}, and models designed specifically to tackle specific tasks such as segmentation in the case of \sam \cite{kirillov2023segany} and \seem \cite{zou2023seem}. 
Notably, exceptional zero-shot capability and strong generalization across a wide range of tasks is reported for each of these models.

Due to constraints related to privacy and ethical considerations, tasks in medical imaging suffer from a shortage of data.
As such, it is an area that can greatly benefit from transfer learning \cite{matsoukas2022makes, matsoukas2023pretrained}. 
It follows that the adoption of foundation models for medical tasks is expected to provide significant advantages as well. 
However, domain shifts between the data that foundation models are pretrained on and the target medical tasks poses a considerable challenge. 
Thus, there is a need to test the effectiveness of these models specifically in this domain. 
Although some attempts have been made to adapt foundation models to the medical domain (\eg adapting \sam for medical segmentation) \cite{qiu2023learnable, ma2023segment, wu2023medical, deng2023segment, shi2023generalist, wang2022medclip, yi2023towards} to the best of our knowledge, a comprehensive evaluation of cutting-edge foundation models for medical image classification is lacking.

In this work, we aim to bridge this gap by conducting a thorough evaluation and analysis of state-of-the-art foundation models in the context of medical image classification, shedding light on their applicability and performance within the medical domain.

The findings of our study are as follows:
\vspace{-2mm}
\begin{itemize}
    \item Not all foundations transfer well to the  medical domain. Some fail to beat the \imagenet-1k baseline. \dinotwo shows performance gains, \seem matches while \sam, \blip and \clip lag behind.
    \vspace{-2mm}
    \item Foundation models require adaptation to downstream tasks. Wherein low-level features demonstrate transferability to medical tasks, while the later layers undergo task-specific adaptation upon unfreezing of the foundation model.
    \vspace{-2mm}
    \item  Appending a \deit classifier to the foundation models results in marginal performance gains. The obtained marginal gains suggest that the high-level features may not effectively serve as inputs to the classifier.
    
    \vspace{-2mm}
\end{itemize}

\noindent
The code to reproduce our experiments can be found at \href{https://github.com/joanaapa/Foundation-Medical}{https://github.com/joanaapa/Foundation-Medical}.

\section{Related Work}
\label{sec:related-work}

Foundation models have sparked tremendous excitement in the community.
In the domain of segmentation tasks, two foundation models have recently gained attention.
The first model, \sam \cite{kirillov2023segany}, incorporates a \vit-based encoder and a lightweight transformer-based decoder, showcasing remarkable zero-shot segmentation capabilities. 
\sam has gained considerable attention in the medical field, despite being a relatively recent publication. 
Several studies have conducted assessments of \sam's performance in medical imaging tasks \emph{out-of the-box}  \cite{cheng2023sam, deng2023segment, shi2023generalist, mohapatra2023brain, zhou2023sam, he2023computervision}. 
Furthermore, researchers have explored  slight modifications of the \sam model to tailor it specifically for medical imaging applications \cite{wu2023medical, qiu2023learnable} or fine-tuned it for specific datasets \cite{ma2023segment}.

\seem \cite{zou2023seem}, another foundation model for segmentation demonstrates excellent performance on numerous benchmarks.
However, at present, there is a dearth of literature exploring \seem's application in the medical domain. 
Despite conducting an exhaustive search, we found no relevant references regarding its usage within the medical domain, making it a compelling candidate for our study.

\dinotwo, introduced by Oquab et al. \cite{oquab_dinov2_2023}, is a \vit-based model that leverages a robust unsupervised pretraining method. 
Building upon \dinos \cite{caron2021emerging}, \dinotwo benefits from training on a large-scale curated dataset, resulting in representations that capture the semantic meaning of images remarkably well. 
This has allowed \dinotwo to demonstrate effective performance across various tasks, including pixel-level and image-level classification.

The \orgclip family of models is pretrained on image-text pairs \cite{radford2021clip}. The combination of language and vision in its representations equips \orgclip with versatile multimodal analysis capabilities, which prove to be powerful in various domains.
The adaptability of \orgclip to the medical domain was demonstrated by Wang et al. \cite{wang2022medclip}. Their study showcased the superior adaptability of \orgclip, surpassing state-of-the-art models specific to radiology. 
Noteworthy studies by Tiu et al. \cite{tiu2022expert}, Zhang et al. \cite{zhang2022contrastive}, and Huang \cite{Huang2021GLoRIAAM} explore similar approaches as \orgclip, showcasing its outstanding zero-shot capabilities. In comparison to its counterparts initialized with ImageNet weights, \orgclip consistently outperforms strong baselines with remarkable data efficiency. \blip \cite{li2022blip} unifies vision-language understanding and generation by employing uni-modal encoders, image-grounded text encoders, and image-grounded text decoders, catering to various image-text tasks. Pre-training involves optimizing Image-Text Contrastive, Image-Text Matching, and Language Modeling objectives. The majority of the discussed studies utilize CNN-based vision backbones initialized with pre-trained weights from \imagenet. Some also incorporate in-domain pretraining on top of \imagenet initialization. 

\section{Methods}
\label{sec:methods}

\addtolength{\tabcolsep}{-3pt}  
\begin{table}[t]
\caption{\textbf{Frozen} foundation models applied to medical tasks.} 
\tiny
\begin{tabular}{ll@{\hskip 5pt}ccccc}
\toprule
\textbf{Classifier} & 
\textbf{Foundation} & &
\textbf{APTOS2019}, $\kappa \uparrow$  &
\textbf{DDSM}, AUC $\uparrow$ &
\textbf{ISIC2019}, Rec. $\uparrow$ & 
\textbf{CheXpert}, AUC $\uparrow$
\\
(Unfrozen) &
(Frozen) & &
$n =$ 3,662  & 
$n =$ 10,239 &
$n =$ 25,333 & 
$n =$ 224,316 \\
\midrule
 & \dino & &
0.878 $\pm$ 0.004 &
0.912 $\pm$ 0.002 &
0.588 $\pm$ 0.015 &
0.738 $\pm$ 0.001  
 
\\
& \ImNet & &
0.864 $\pm$ 0.007 &
0.908 $\pm$ 0.001 &
0.513 $\pm$ 0.015 &
0.719 $\pm$ 0.001  
\\

& \resnetOneFiftyTwo  & &
0.824 $\pm$ 0.003 &
0.883 $\pm$ 0.001 &
0.461 $\pm$ 0.027 &
0.712 $\pm$ 0.000 
\\
\cmidrule{1-7}
\multirow{4}{*}{Linear} 
& \sam & &
0.873 $\pm$ 0.010 &
0.916 $\pm$ 0.004 &
0.402 $\pm$ 0.014 &
0.724 $\pm$ 0.001
\\
& \seem & &
0.852 $\pm$ 0.003 &
0.891 $\pm$ 0.001 &
0.527 $\pm$ 0.007  &
0.731 $\pm$ 0.001
\\
& \clip & &
0.857 $\pm$ 0.003 &
0.897 $\pm$ 0.002 &
0.489 $\pm$ 0.010 &
0.702 $\pm$ 0.001
\\
& \dinotwo & &
0.881 $\pm$ 0.002 &
0.905 $\pm$ 0.001 &
0.569 $\pm$ 0.012 &
0.722 $\pm$ 0.000 
\\
& \blip & &
0.818 $\pm$ 0.014 &
0.845 $\pm$ 0.005 &
0.416 $\pm$ 0.021 &
0.647 $\pm$ 0.000 

\\[0.5em]
\cmidrule{1-7}
\multirow{4}{*}{\deitbase} 
& \sam & &
0.890 $\pm$ 0.005 &
0.940 $\pm$ 0.007  &
0.740 $\pm$ 0.015 &
0.788 $\pm$ 0.003
\\
& \seem & &
0.887 $\pm$ 0.002 &
0.925 $\pm$ 0.006 &
0.747 $\pm$ 0.028 &
0.777 $\pm$ 0.001
\\
& \clip & &
0.903 $\pm$ 0.006 &
0.948 $\pm$ 0.004 &
0.748 $\pm$ 0.010 &
0.790 $\pm$ 0.001
\\
& \dinotwo & &
0.901 $\pm$ 0.005 &
0.945 $\pm$ 0.008 &
0.790 $\pm$ 0.010 &
0.798 $\pm$ 0.001  
\\
& \blip & &
0.895 $\pm$ 0.009 &
0.945 $\pm$ 0.004 &
0.763 $\pm$ 0.011 &
0.785 $\pm$ 0.000 
\\[0.5em]
\bottomrule
\end{tabular}

\label{tab:frozen_table}
\vspace{-1mm}
\end{table}
\addtolength{\tabcolsep}{3pt}

The primary objective of this study is to evaluate the efficacy of vision foundation models in the context of medical image classification.
Specifically, we aim to assess the transferability of features generated by these models to the medical domain and identify the best utilization strategies. 
The foundation models we consider were trained with different data and different objectives, so it stands to reason that some models will be better suited to transfer features for medical tasks than others.
We conducted a series of experiments employing two distinct approaches: using the foundation model as a standalone model (with a linear prediction head) and stacking it with a model referred to as the "Classifier model". 
For consistency, \deit \cite{touvron2021training}, a member of the transformer family, was used as the classifier.

\subsection{Datasets}
To ensure a comprehensive assessment of feature transferability to the medical domain, we carefully chose four publicly available and well-established datasets. These, encompass a wide range of imaging modalities, color scales, and dataset sizes, providing diversity in the evaluation scenarios. For each dataset, we employed specific evaluation metrics tailored to the nature of the classification task, enabling a detailed analysis of model performance.

\begin{itemize}
    \item \textbf{APTOS 2019} \cite{aptos2019} - The dataset comprises a collection of 3,662 high-resolution images related to diabetic retinopathy, categorized into five severity classes. The evaluation of model performance on this dataset is measured using the quadratic Cohen kappa metric.
    \item \textbf{CBIS-DDSM} \cite{ddsm} - The dataset used in this study consists of 10,239 mammography images for the detection of anomaly masses. The model evaluation on this dataset was conducted using the ROC-AUC metric.
    \item \textbf{ISIC 2019} \cite{isic} - This dataset  comprises a set of 25,331 dermoscopic images. These images were used for the classification of skin lesions into 9 distinct diagnostic categories. The evaluation of model performance on this dataset was assessed using the recall metric.
    \item \textbf{CHEXPERT} \cite{irvin2019chexpert} - This dataset contains a series of 224,316 chest X-rays containing 14 different diagnostic observations,  encompassing a range of conditions and abnormalities. The model performance was evaluated using the ROC-AUC metric.
\end{itemize}

\addtolength{\tabcolsep}{-3pt}  
\begin{table}[t]
\caption{\textbf{Unfrozen} foundation models applied to medical tasks.
}
\tiny
\begin{tabular}{ll@{\hskip 5pt}ccccc}
\toprule
\textbf{Classifier} & 
\textbf{Foundation} & &
\textbf{APTOS2019}, $\kappa \uparrow$  &
\textbf{DDSM}, AUC $\uparrow$ &
\textbf{ISIC2019}, Rec. $\uparrow$ & 
\textbf{CheXpert}, AUC $\uparrow$
\\
(Unfrozen) &
(Unfrozen) & &
$n =$ 3,662  & 
$n =$ 10,239 &
$n =$ 25,333 & 
$n =$ 224,316 \\
\midrule
 & \dino & &
0.886 $\pm$ 0.004 &
0.958 $\pm$ 0.004 &
0.775 $\pm$ 0.007  &
0.794 $\pm$ 0.001 
\\
& \ImNet & &
0.904 $\pm$ 0.006 &
0.960 $\pm$ 0.003 &
0.823 $\pm$ 0.008 &
0.797 $\pm$ 0.002  
\\

& \resnetOneFiftyTwo & &
0.899 $\pm$ 0.002 &
0.960 $\pm$ 0.003 &
0.817 $\pm$ 0.007 &
0.807 $\pm$ 0.000 
\\
\cmidrule{1-7}
\multirow{4}{*}{Linear} 
& \sam & &
0.894 $\pm$ 0.007 &
0.950 $\pm$ 0.006 &
0.798 $\pm$ 0.010 &
0.801 $\pm$ 0.001
\\
& \seem & &
0.902 $\pm$ 0.003 &
0.958 $\pm$ 0.003 &
0.835 $\pm$ 0.009 &
0.808 $\pm$ 0.001  
\\
& \clip & &
0.903  $\pm$ 0.013 &
0.945  $\pm$ 0.012 &
0.818  $\pm$ 0.008 &
0.806  $\pm$ 0.001
\\
& \dinotwo & &
\textbf{0.909} $\pm$ 0.009 &
\textbf{0.966} $\pm$ 0.003 &
\textbf{0.859} $\pm$ 0.007 &
\textbf{0.812} $\pm$ 0.001 
\\

& \blip & &
0.891 $\pm$ 0.006 &
0.961 $\pm$ 0.003 &
0.798 $\pm$ 0.016 &
0.797 $\pm$ 0.000 
\\[0.5em]
\cmidrule{1-7}
\multirow{4}{*}{\deitbase} 
& \sam & &
0.890 $\pm$ 0.005 &
0.951 $\pm$ 0.004 &
0.778 $\pm$ 0.013 &
0.796 $\pm$ 0.002   
\\
& \seem & &
0.901 $\pm$ 0.007 &
0.946 $\pm$ 0.002 &
0.808 $\pm$ 0.022 &
0.802 $\pm$ 0.001
\\
& \clip & &
0.907 $\pm$ 0.011 &
0.955 $\pm$ 0.010 &
0.809 $\pm$ 0.013 &
0.805 $\pm$ 0.001 
\\
& \dinotwo & &
0.904 $\pm$ 0.007 &
\textbf{0.966} $\pm$ 0.004 &
0.822 $\pm$ 0.011 &
\textbf{0.812} $\pm$ 0.001
\\
& \blip & &
0.897 $\pm$ 0.001 &
0.955 $\pm$ 0.002 &
0.806 $\pm$ 0.016 &
0.799 $\pm$ 0.000 
\\[0.5em]
\bottomrule
\end{tabular}
\label{tab:main_table}
\end{table}
\addtolength{\tabcolsep}{3pt}

\subsection{Foundation models}
\label{sec:foundation}
The foundation models chosen for this study are all based on the transformer architecture. 
In order to maintain fairness in our comparisons, we specifically selected versions of these models that were as closely matched in size as feasible. 
This approach ensures that any performance differences observed can be attributed to the specific architectural variances and not to significant differences in model size or complexity.
The foundation models we consider in this study are:

\begin{itemize}
    \item \textbf{\sam} (88.8M parameters) \cite{kirillov2023segany} - \sam is an encoder-decoder architecture specifically designed for promptable segmentation tasks. It has been trained on a dataset consisting of 11 million high-resolution images along with $1$ billion associated masks. We utilized the image encoder of \sam as the foundation model. When fine-tuning \sam, we use a resolution of $224$ which deviates from the original training resolution 
    \item \textbf{\seem} (29.9M parameters) \cite{zou2023seem} - \seem employs a generic encoder-decoder architecture comprised of separate encoders for vision and text, followed by a decoder for mask generation which works on the joint embedding space of vision and text. A combination of \coco \cite{lin2015microsoft} and \refcoco dataset \cite{yu2016modeling} were utilized for training. Our feature extractor is the Focal transformer \cite{yang2022focal} vision backbone. 
    
    \item \textbf{\dinotwo} (86.5M parameters) \cite{oquab_dinov2_2023} - This all-purpose \vit model is pretrained using a discriminative self-supervised method. The pretraining dataset ($142$ million images) is a curation of $24$ publicly available natural image datasets for classification, retrieval and segmentation tasks.  We use the \vit-B model, which is a distilled version of their \vit-H architecture. 
    
    \item \textbf{\clip} (86.2M parameters) \cite{radford2021clip} - A versatile \vit model, employing customized text-guided pretraining on a subset of the \laion dataset \cite{schuhmann2022laion5b}. The subset \laiontwob contains $2$ billion image-caption(in english) pairs obtained through web scraping. For our experiments, we utilize the base version of this model.

    \item \textbf{\blip} (85.7M parameters) \cite{li2022blip} - Utilizing vision transformers and BERT-based encoders, BLIP integrates visual and textual information through cross-attention and specialized tokens. Trained on  $142$ million images curated from datasets ranging from \laionfour to \coco. Our feature extractor is the \vit-B vision backbone.
    
\begin{figure}[t]
\centering
\scriptsize
\begin{tabular}{@{}c@{}c@{}}

\hspace{5mm}\aptos &
\hspace{5mm}\ddsm 
\\

\includegraphics[width=0.5\linewidth]{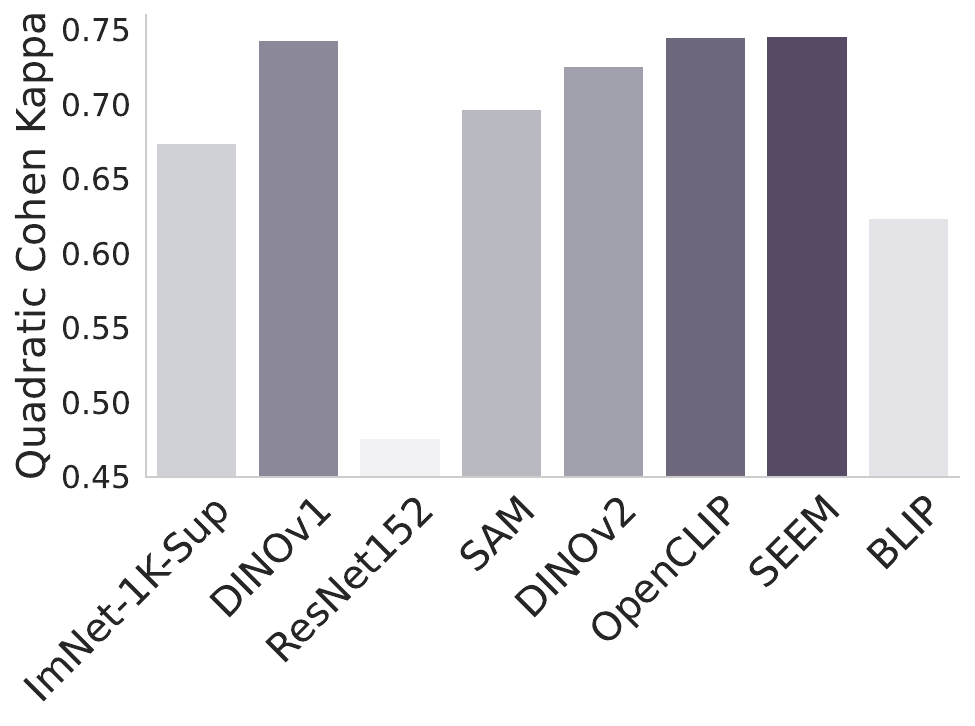} &
\includegraphics[width=0.5\linewidth]{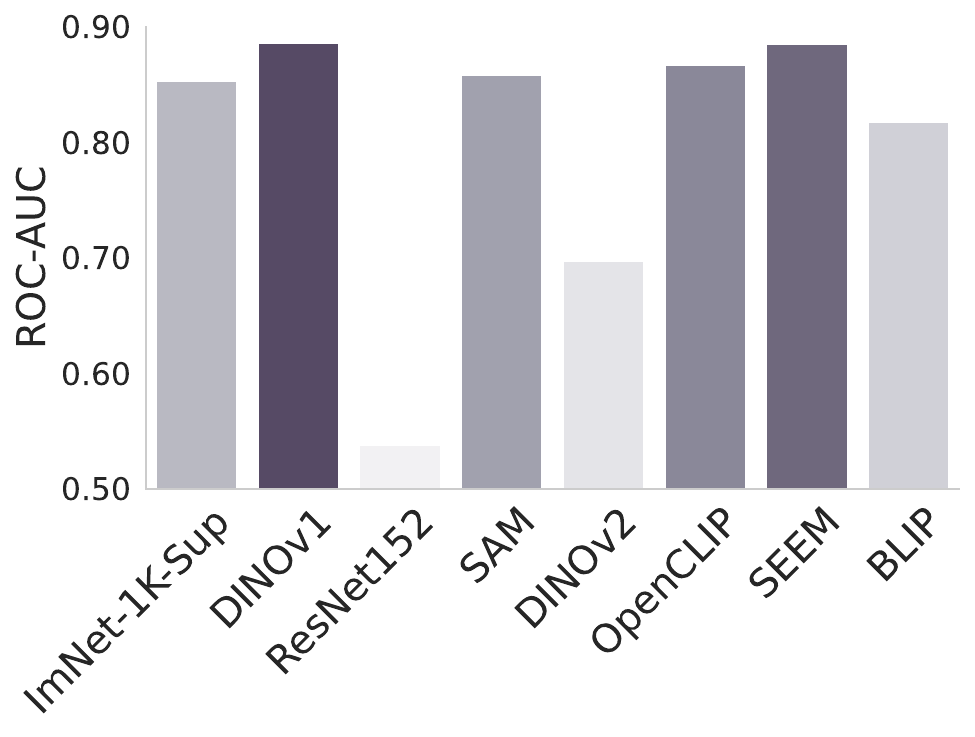} \\
\vspace{-1mm}
\hspace{5mm}\chexpert &
\hspace{5mm}\isic
\\
\includegraphics[width=0.5\linewidth]{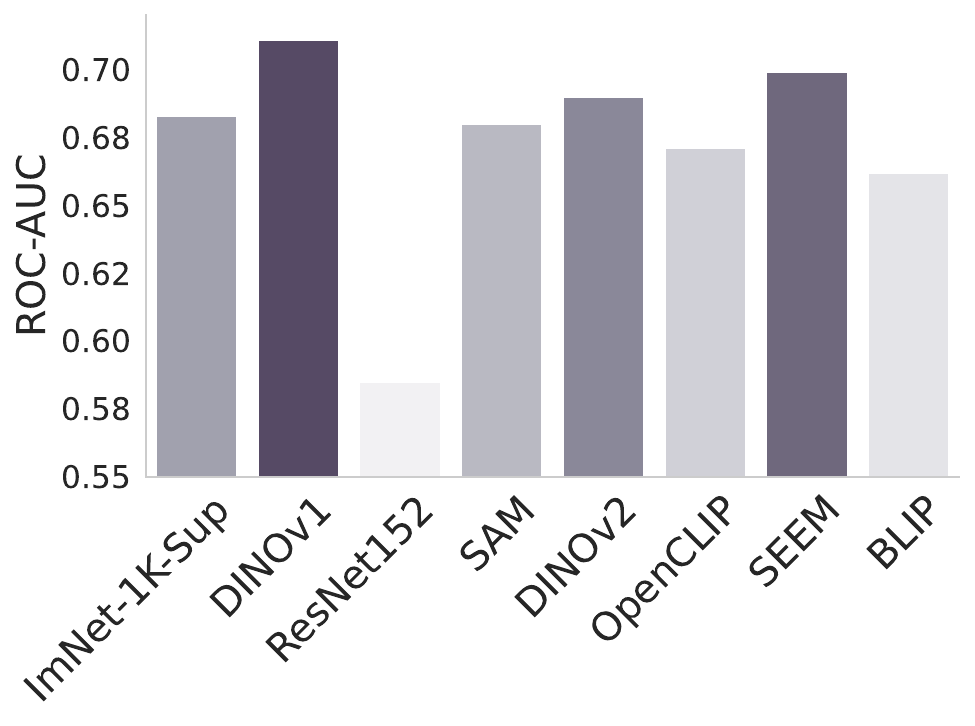}&
\includegraphics[width=0.5\linewidth]{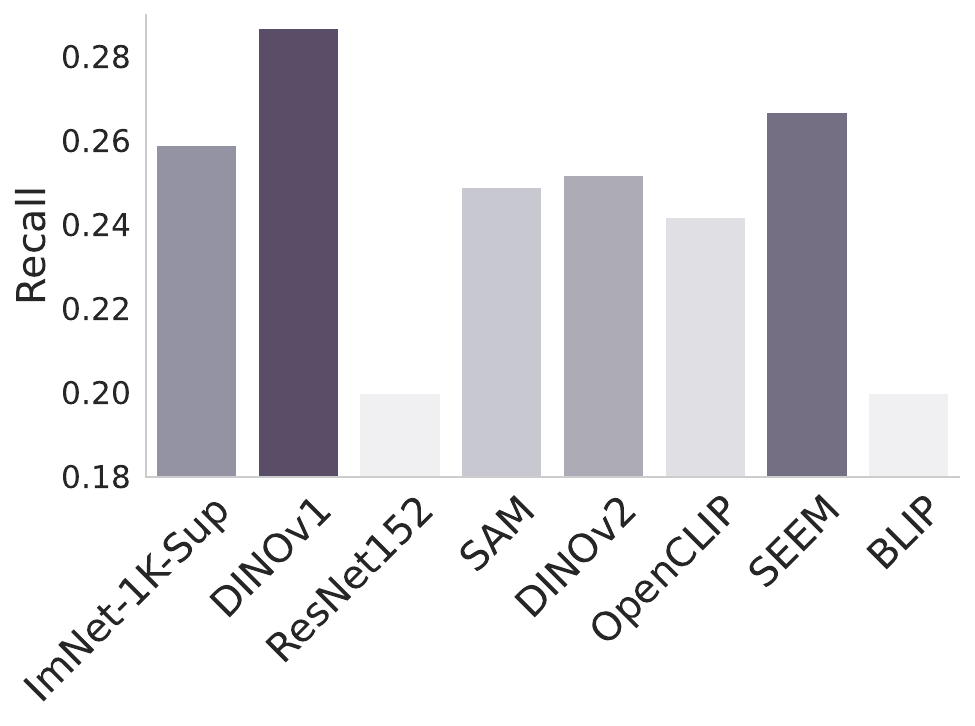} 
\\

\end{tabular} 
\vspace{-4mm}
   \caption{ $k$-NN evaluation performance of different foundation and baseline models.}
\vspace{-4mm}   
\label{fig:knn}
\end{figure}

\end{itemize}
In addition to these foundation models, we consider as a baselines models trained on \imagenet-1k (ILSVRC-2012) \cite{russakovsky2015imagenet}
\begin{itemize}
    \item \textbf{\ImNet} (86.6M parameters) - The standard practice for transfer learning to medical classification tasks is to use \imagenet-1k pretrained networks \cite{matsoukas2022makes} for initialization. In this work we use \deitbase \cite{touvron2021training} pretrained in a supervised manner on \imagenet-1k as our baseline.

    \item \textbf{\dino}  (86.6M parameters) \cite{caron2021emerging} - We also include in our comparisons this general-purpose \vit model, which is pretrained using an older self-supervised method compared to \dinotwo, but on \imagenet-1k \cite{russakovsky2015imagenet}. In this work we use the base model, which is based on the \deitbase architecture.

    \item \textbf{\resnet}  (60.4M parameters) \cite{he2015deep} - We have also included this general purpose \cnn backbone, which is pretrained on \imagenet-1k. We use \resnetOneFiftyTwo to match the complexity of other baselines.

\end{itemize}

\subsection{Implementation Details}
\label{sec:implementation}
Prior to training, all images were downsampled to a resolution of $256\times256$, followed by a center crop to obtain a final size of $224\times224$. 
A set of standard augmentations was applied to the training images. 
These augmentations included normalization, vertical and horizontal flips, color jitter (adjusting brightness, contrast, saturation, and hue), and random resize crops. 
Each training set was divided into train (80\%), validation (10\%) and test (10\%). 
In the case of APTOS, which is of a smaller size, the split was adjusted to 70\%, 15\% and 15\%, respectively. 
Some foundation models like \sam were trained with different resolutions.
To handle this, we use bicubic interpolation to align the position embeddings with the image scales and ensure compatibility \cite{dosovitskiy2021image}.

\begin{figure}[t]
\centering
\scriptsize
\begin{tabular}{@{}l}

\hspace{5mm} Train time - Unfrozen 
\hspace{6mm} Train time - Frozen 
\hspace{9mm} Inference time
\\
\includegraphics[width=\linewidth]{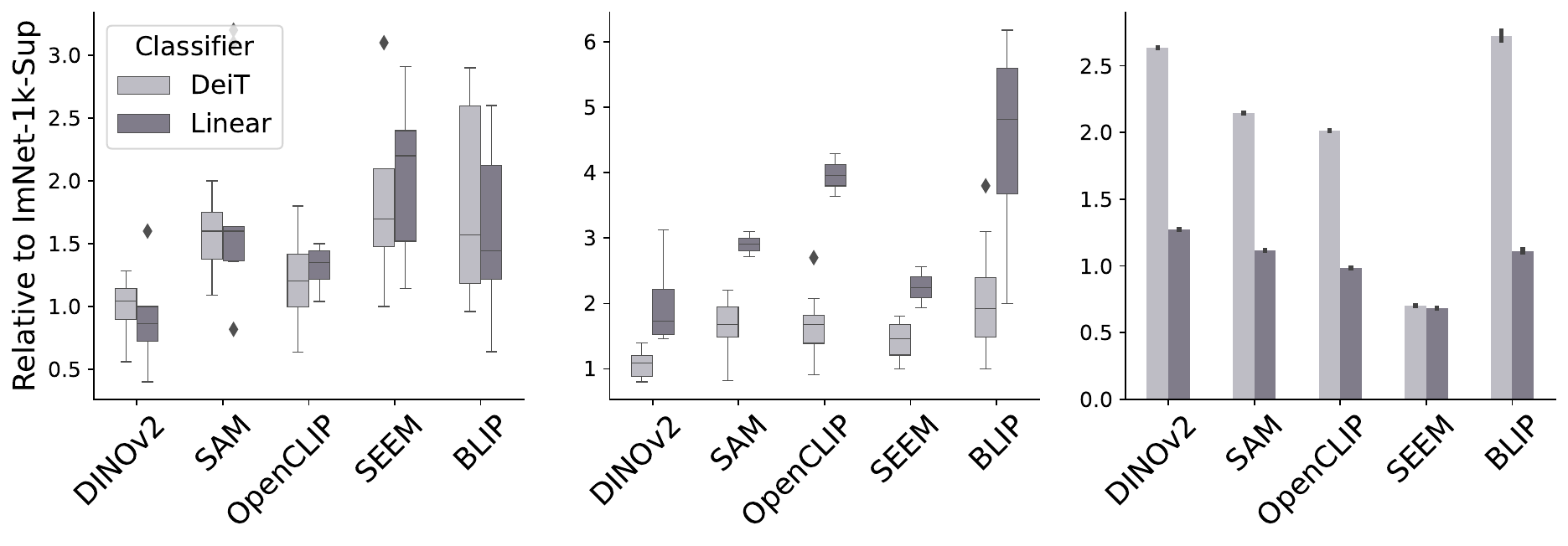}
\end{tabular} 
\vspace{-4mm}
\caption{Training and inference times.}
\label{fig:training_test}
\vspace{-4mm}
\end{figure}

In addition to the training resolution, the foundation model's architectures exhibit significant differences that necessitate adaptation for the medical tasks at hand. 
Specifically, when considering the \seem and \sam adaptations of vision transformer architectures, the absence of a \emph{[cls]} token requires us to employ global average-pooling of the patch representations obtained from the final layer in order to attach a linear head or perform a $k$-NN evaluation. 
In case of \clip, \blip and \dinotwo, we can make use of the original \emph{[cls]} token without modification. 
In all cases, integrating a classifier model on top of the foundation model involves extracting patch representations from the foundation models, adding a projection layer, bypassing the patch embedding layer of the stacked classifier, and transmitting the patch representations to the classifier for the fine-tuning process. The appended Classifier's \emph{[cls]} token is used for the classification task.

The models were fine-tuned using supervised learning on each medical dataset with the AdamW optimizer \cite{loshchilov2019decoupled}, employing a linear learning rate warmup strategy. 
To determine the optimal base learning rate, a search was conducted across four values ranging from $10^{-5}$ to $10^{-3}$. 
As the saturation point was reached, the learning rate was decreased by a factor of 10. 
A weight decay of $10^{-5}$ was applied, and this entire process was repeated five times to account for performance variance. 
The evaluation of the models was based on selecting the best-performing checkpoint, identified through analysis on the validation set.

\begin{figure}[t]
\centering
\scriptsize
\begin{tabular}{@{}c@{}c@{}}
\hspace{5mm}\aptos &
\hspace{5mm}\ddsm 

\\

\includegraphics[width=0.5\linewidth]{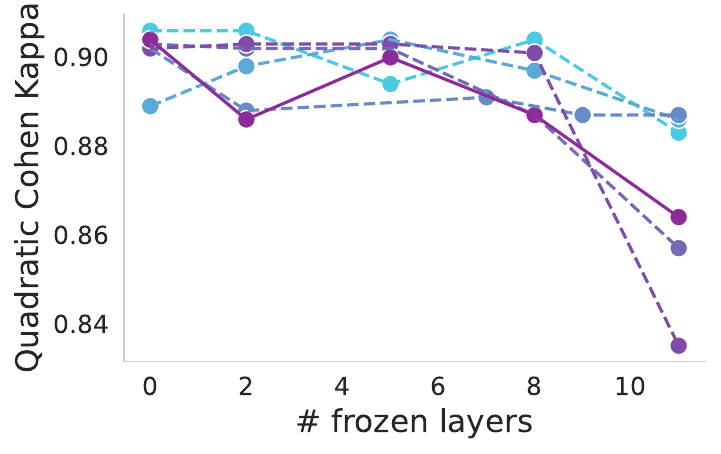} &
\includegraphics[width=0.5\linewidth]{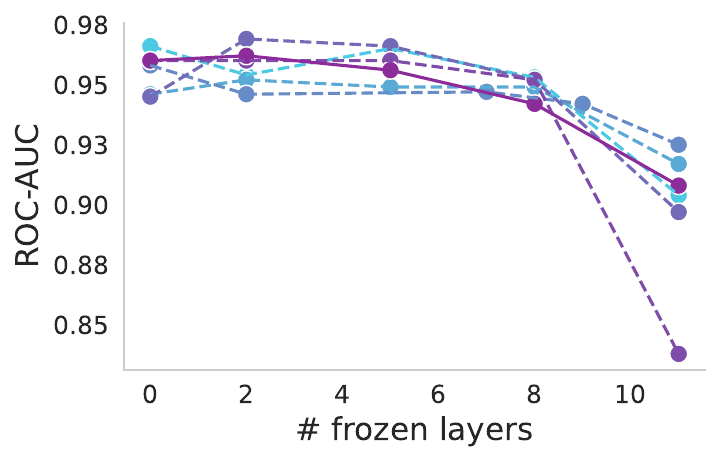} \\
\hspace{5mm}\chexpert  &
\hspace{5mm}\isic
\\
\includegraphics[width=0.5\linewidth]{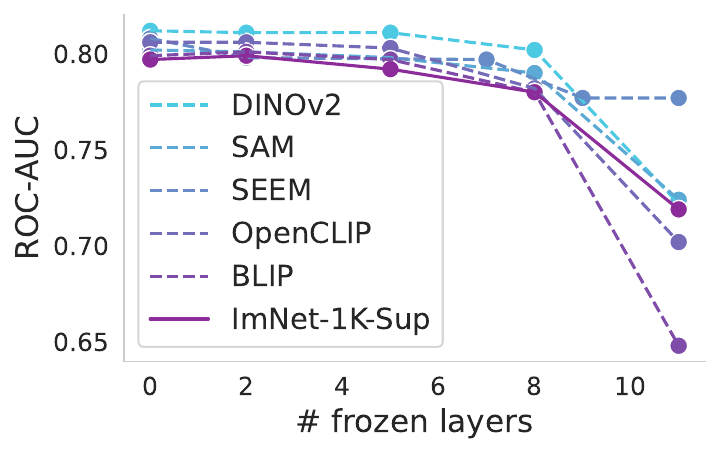} &
\includegraphics[width=0.5\linewidth]{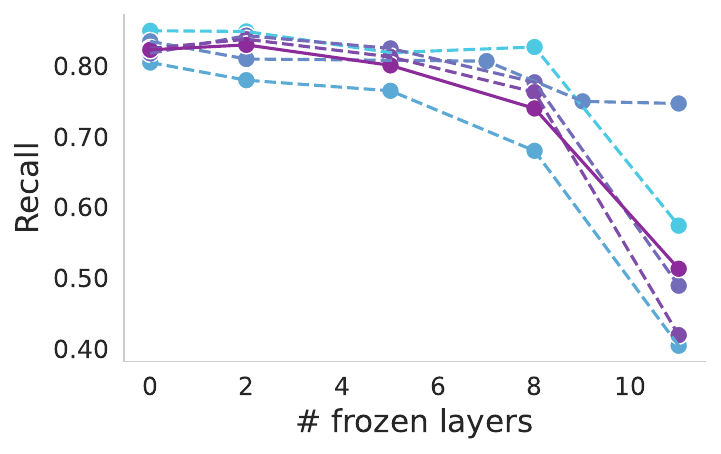} \\
\end{tabular}
\vspace{-2mm}
\caption{Impact of gradually freezing layers of the foundation model on performance for medical tasks (using a linear head).} 
\label{fig:partially_freezing}
\vspace{-4mm}
\end{figure}

\subsection{Evaluation protocol}
\label{sec:evaluation}
We employed a two-step approach to assess the performance of the foundation models. 
Initially, we kept the foundation models frozen and performed transfer learning by fine-tuning a linear classifier head on top. 
This enabled us to evaluate the adapted model's performance in comparison to the baseline (\ImNet). 
We conducted this evaluation across multiple medical datasets to gauge the model's efficacy.
Subsequently, we appended a transformer classifier (\deitbase) on top of the frozen foundation. 
We fine-tuned this combined architecture on the medical tasks and assessed its performance. 
To gain further insights, we repeated the aforementioned procedure while unfreezing the weights of the foundation model. 
This allowed the foundation model to adapt its features during fine-tuning, both with the linear head and with the classifier.
To summarize, we considered four scenarios:
\begin{enumerate}
    \item \textbf{Frozen with linear head.} Transfer learning to the medical tasks is done while keeping the foundation model frozen, only a linear head is fine-tuned to the task.
    \item \textbf{Frozen with appended classifier.} Transfer learning to the medical tasks is done while keeping the foundation model frozen, a \deitbase classifier is appended and fine-tuned for the task.
    \item \textbf{Unfrozen with linear head.} Both the foundation model and a linear head are fine-tuned to the task.
    \item \textbf{Unfrozen with appended classifier.} Both the foundation model and an appended \deitbase classifier are fine-tuned for the task.
\end{enumerate}

\begin{figure}[t]
\centering
\scriptsize
\begin{tabular}{@{}c@{}c@{}}
\hspace{5mm}\aptos - Unfrozen foundation &
\hspace{5mm}\ddsm  - Unfrozen foundation

\\

\includegraphics[width=0.5\linewidth]{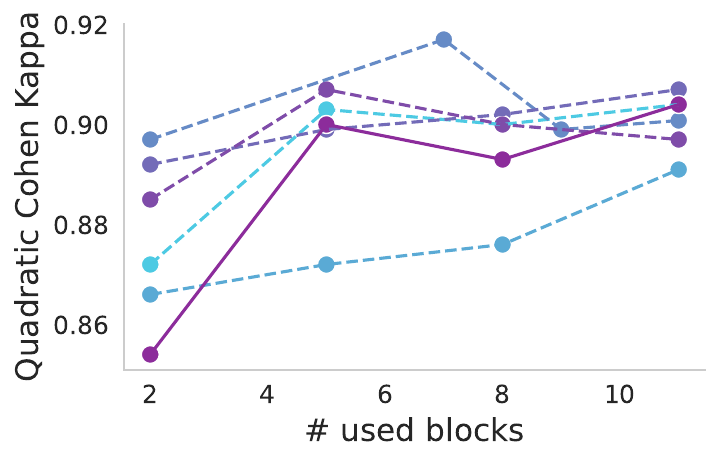} &
\includegraphics[width=0.5\linewidth]{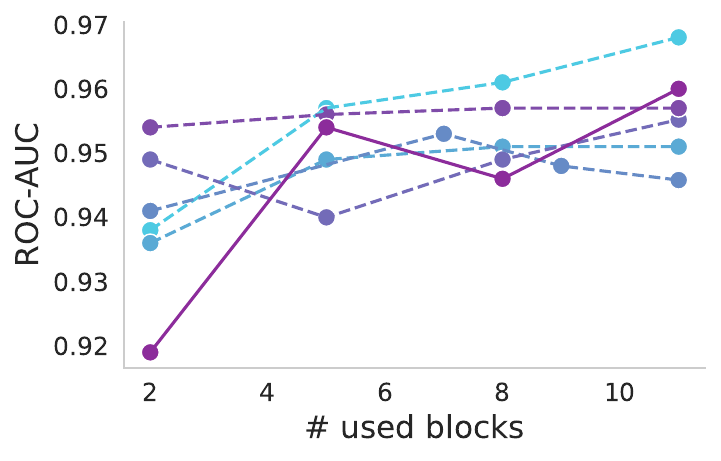} \\
\hspace{5mm}\aptos - Frozen foundation &
\hspace{5mm}\ddsm - Frozen foundation
\\
\includegraphics[width=0.5\linewidth]{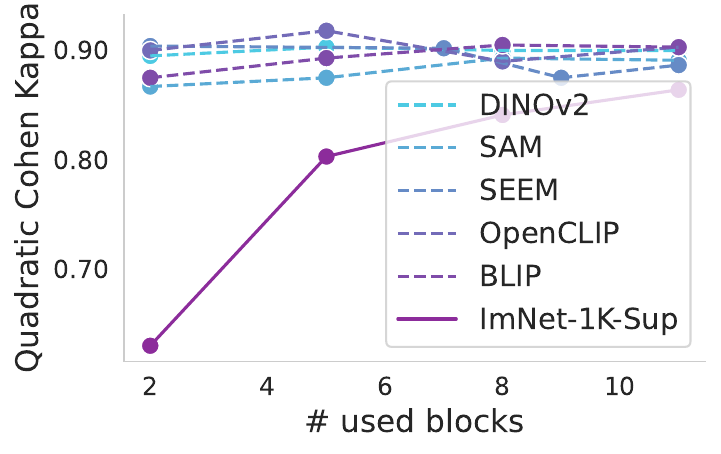} &
\includegraphics[width=0.5\linewidth]{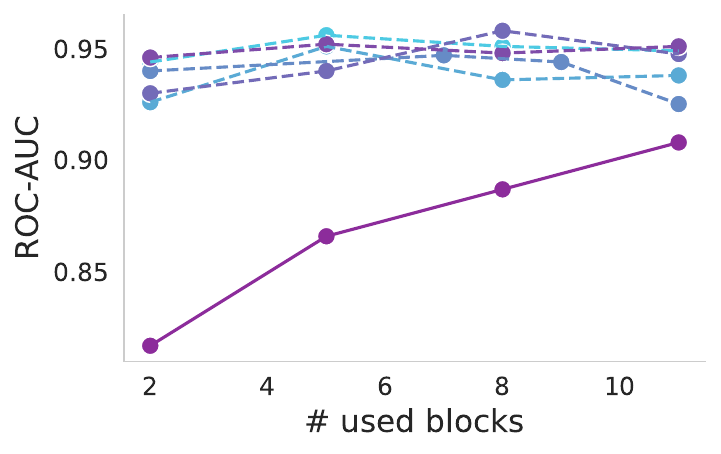} \\

\end{tabular} 
\vspace{-2mm}
\caption{Model performance when using different subsets of foundation layers in conjunction with a \deit classifier.} 
\label{fig:dropping}
\vspace{-4mm}
\end{figure}

In the above setups we evaluated the final performance of the fine-tuned models on the medical tasks.
We also consider the generalization capability and robustness to class separability of the features, which can be assessed using a \emph{k}-Nearest Neighbour evaluation on the features extracted by the foundation model.
\begin{enumerate}[label={\arabic*.}, start=5]
    \item \textbf{$k$-NN evaluation.} We examine the applicability of the foundation models features for the medical tasks using $k$-Nearest Neighbors.
\end{enumerate}

Furthermore we evaluate the impact of model complexity and resolution on the performance as well as training and inference times
\begin{enumerate}[label={\arabic*.}, start=6] 
    \item \textbf{Model size.} We select the top-performing model and evaluation protocol, then examine various available models of differing sizes. 
    \item \textbf{Image resolution.} We upscale training resolution to $1024 \times 1024$ and evaluate using the best-identified scenario.
    \item \textbf{Train and inference times.} We define training time as the number of iterations until the validation metrics reaches maximum performance (normalized across datasets). Inference time is calculated as the average time to classify a single instance. Both are reported relative to the baseline \ImNet.
\end{enumerate}

Finally, to investigate how internal representations within the foundation models adapt to the medical tasks, we employ Centered Kernel Alignment (CKA) to compare models before and after fine-tuning.
\begin{enumerate}[label={\arabic*.}, start=9]
    \item \textbf{Centered Kernel Alignment (CKA).} We employ Centered Kernel Alignment (CKA) \cite{cortes2014algorithms} to analyze the adaptation of internal representations within the foundation models to the medical tasks. CKA is a metric that measures the similarity between layers in different neural networks. This allows us to gain insights into the specific layers that undergo changes during the fine-tuning process. To visualize and quantify these changes, we compute a CKA heat map comparing the pre- and post-fine-tuned versions of the foundation models. 
\end{enumerate}

\section{Experiments}
\label{sec:experiments}

\begin{figure}[t]
\centering
\scriptsize
\begin{tabular}{@{}l@{}l@{}l@{}}
\hspace{8mm}\aptos &
\hspace{11mm}\ddsm &
\hspace{10mm}\isic

\\

\includegraphics[width=0.35\linewidth]{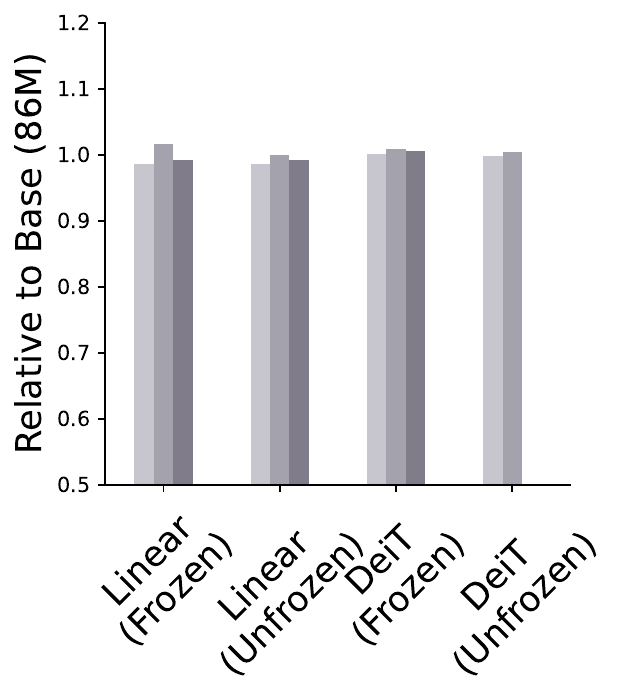} &
\includegraphics[width=0.32\linewidth] {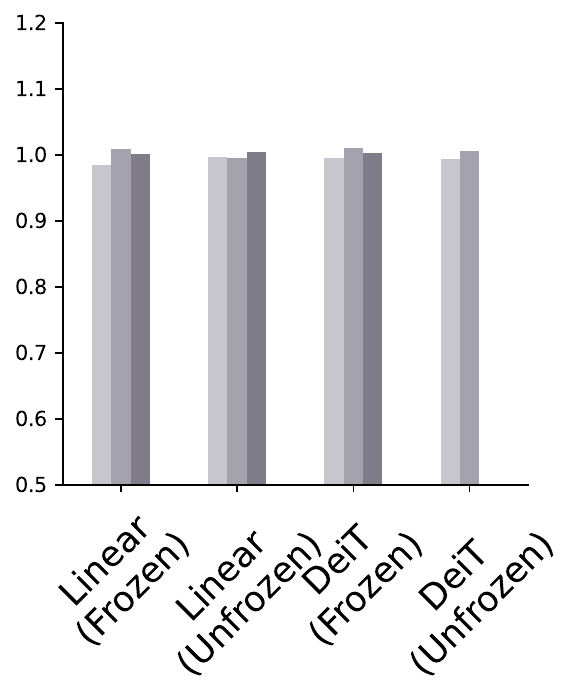} &

\includegraphics[width=0.32\linewidth]{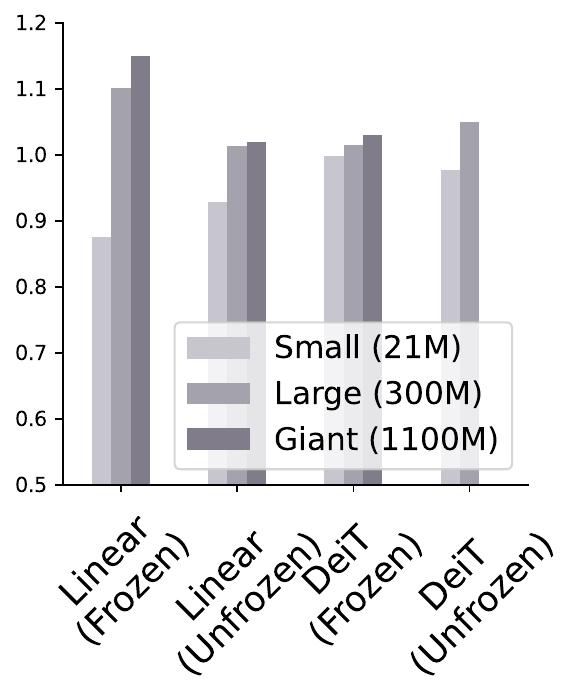} \\

\end{tabular} 
\vspace{-3mm}
\caption{Performance of \dinotwo with different model complexities relative to the Base model.} 
\label{fig:sizes}
\vspace{-4mm}
\end{figure}

In this section, we present the findings from the experiments outlined in Sections \ref{sec:foundation}, \ref{sec:implementation}, and \ref{sec:evaluation}.

\subsection{Foundational Features for Medical Tasks}
\label{sec:usefulness}
To measure the direct transferability of the features produced by the foundation models for medical tasks, we froze the foundation model and appended a linear head (1) or a classifier (2) as described in Section \ref{sec:evaluation}. The results of this experiment are provided in Table \ref{tab:frozen_table}.
When compared to the unfrozen fine-tuning, this approach was generally inferior.

\noindent\textbf{Frozen and unfrozen foundations.} We also measure the ability of foundation models to adapt to medical tasks by unfreezing their weights during fine-tuning, both with a linear head (3) and an appended classifier (4) as described in Section \ref{sec:evaluation}.
The results of this experiment are provided in Table \ref{tab:main_table}. 
Here, it can be seen that \dinotwo consistently provides significant boosts in performance when transferred to medical tasks. 
The results with linear classifiers in Table \ref{tab:main_table} demonstrate that \dinotwo  outperforms the baseline across all datasets, yielding performance gains of up to 3.2\%.
The performance achieved on the \aptos and \isic highlights the potential for significant improvements even when dealing with small datasets. In fact, \dinotwo seems to outperform even self-supervised \resnets, pretrained in-domain, as it can be seen in Appendix \ref{apx:ssl}.

\addtolength{\tabcolsep}{-3pt}  
\begin{table}[t]
\caption{\textbf{Foundation models} trained at $1024 \times 1024$ resolution. Percentage gain relative to low resolution indicated in brackets. }
\resizebox{8.4cm}{!}{
\begin{tabular}{lc@{\hskip 5pt}cccc}

\toprule
\textbf{Foundation} & 
\textbf{APTOS2019}, $\kappa \uparrow$  &
\textbf{DDSM}, AUC $\uparrow$ &
\textbf{ISIC2019}, Rec. $\uparrow$ & 
\textbf{CheXpert}, AUC $\uparrow$
\\
(UnFrozen) &
$n =$ 3,662  & 
$n =$ 10,239 &
$n =$ 25,333 & 
$n =$ 224,316 \\
\midrule
\sam & 
0.908 (1.4) &
0.984 (3.4) &
0.804 (0.6) &
0.820 (1.9)
\\
\seem  &
0.911 (0.9) &
0.967 (0.2) &
0.868 (3.3) &
0.821 (1.3)
\\
\clip  &
0.904  (0.1) &
0.980  (3.5) &
0.852  (3.4) &
0.813  (0.7)
\\
\dinotwo &
0.917 (0.8) &
0.983 (1.7) &
0.895 (3.6) &
0.825 (1.3)
\\
\blip &
0.909 (1.2) &
0.984 (2.9) &
0.839 (3.3) &
0.813 (1.4)
\\
\bottomrule
\end{tabular}

}
\label{tab:resolution}
\vspace{-3mm}
\end{table}
\addtolength{\tabcolsep}{3pt}

\noindent\textbf{$k$-NN evaluation.} 
In Figure \ref{fig:knn} we examine the adaptability of the features from the foundation models using $k$-Nearest Neighbors.
Without any fine-tuning or supervision, this tests how well the foundation model can separate the classes in the task at hand.
The results are largely in line with the findings of Tables \ref{tab:frozen_table} and \ref{tab:main_table}, but surprisingly \dino performs the best overall, despite its poor performance when fine-tuned for the task.
Another surprising observation is the consistently poor performance of \ImNet and \resnet. Their raw features perform poorly measured by $k$-NN but they adapt to the task through fine-tuning more readily than \sam and \clip.

\noindent\textbf{Progressive freezing.} 
It is known that different layers capture different image characteristics, starting from low level features in the first layers to high level at the last ones.
Above, we only considered two scenarios, which were to use the features that the foundation provided out-of-the box or completely adapting the foundation during training. 
Here, we investigate the suitability of the features produced by each layer of the foundation models.
Figure \ref{fig:partially_freezing} illustrates the performance of foundation models when trained with a linear classifier by progressively freezing blocks. 
A consistent trend can be observed across almost all cases,  including the baseline. 
Freezing up to the eighth block leads to a marginal decrease in performance, but beyond that point, there is a significant decline, which is particularly notable for \isic.
This may indicate a point where high-level foundational features become too specialized to the pretraining task, and unsuited for medical tasks.
The only exception is \seem not exhibiting this significant drop in performance.

\begin{figure}[!t]
\begin{center}
   \includegraphics[width=\linewidth]{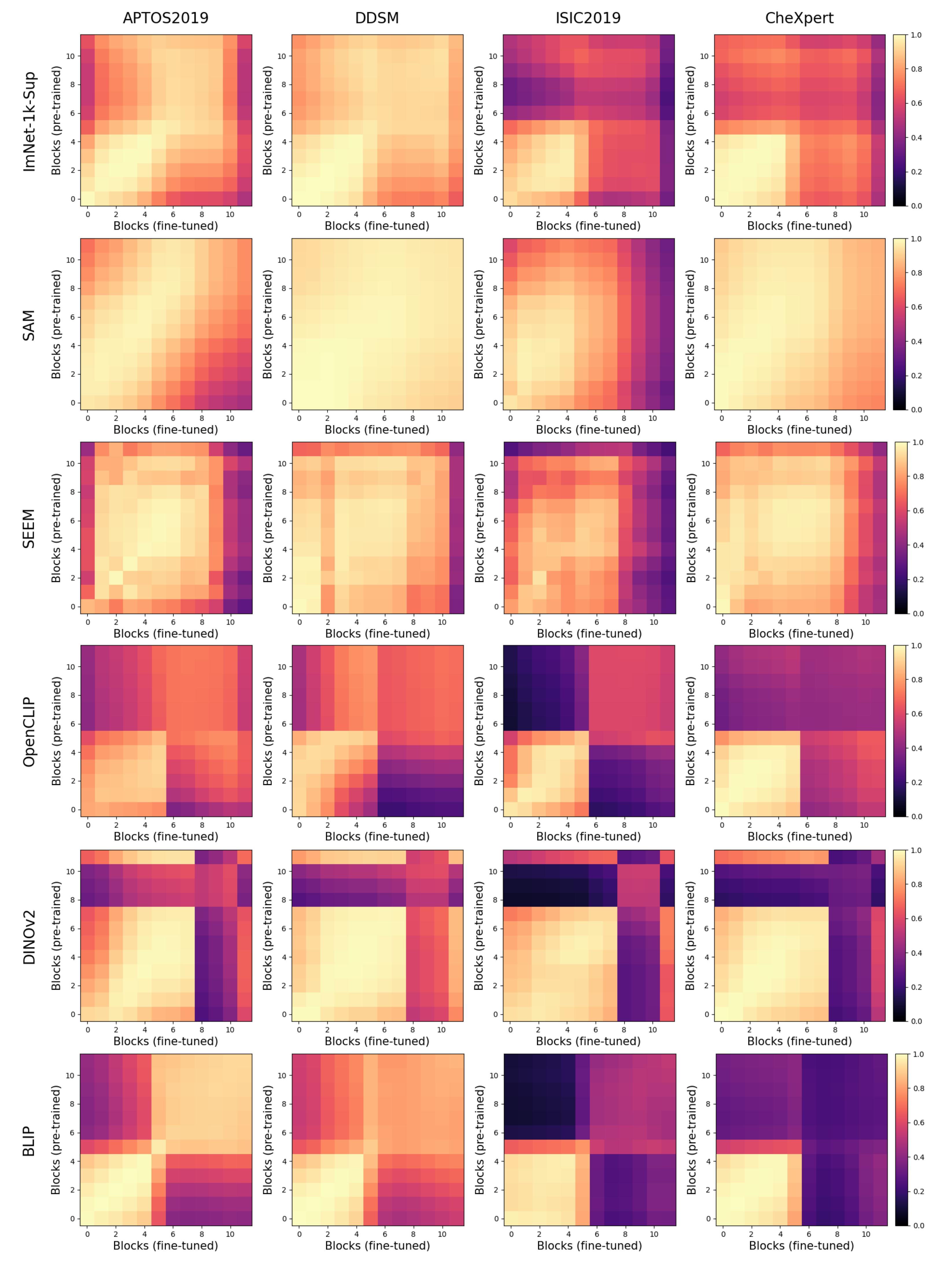}
\end{center}
\vspace{-4mm}
   \caption{Layerwise CKA heatmaps pre-trained and fine-tuned foundation models using a linear classifier. For reference, the baseline \ImNet is included.}
\label{fig:fietuned_representations}
\vspace{-5mm}
\end{figure}

\begin{figure}[t]
\centering
\scriptsize
\begin{tabular}{@{}l}
\hspace{23mm}\aptos  
\hspace{21mm}\ddsm  
\\
\includegraphics[width=\linewidth]{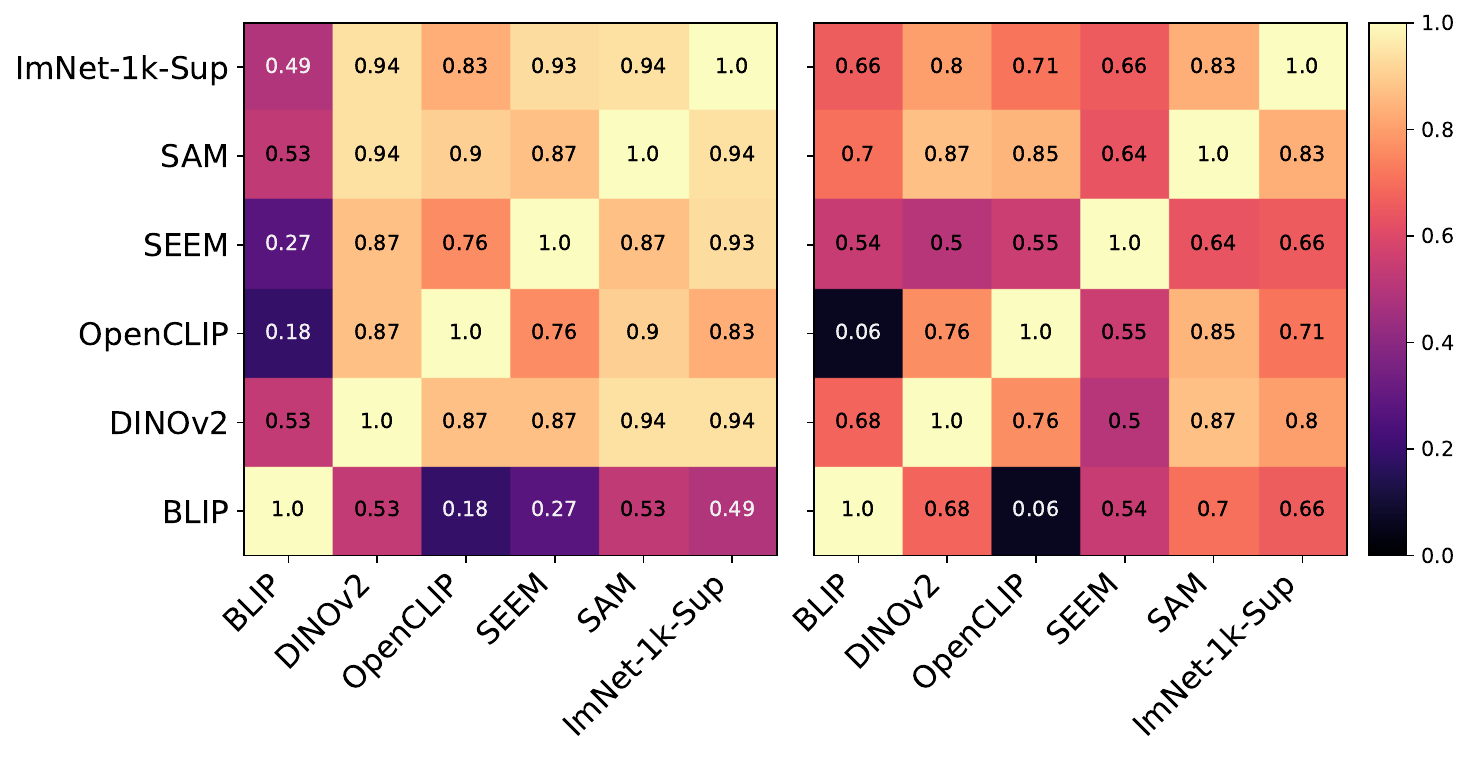} 
\\
\hspace{24mm}\chexpert  
\hspace{23mm}\isic  
\\
\includegraphics[width=\linewidth]{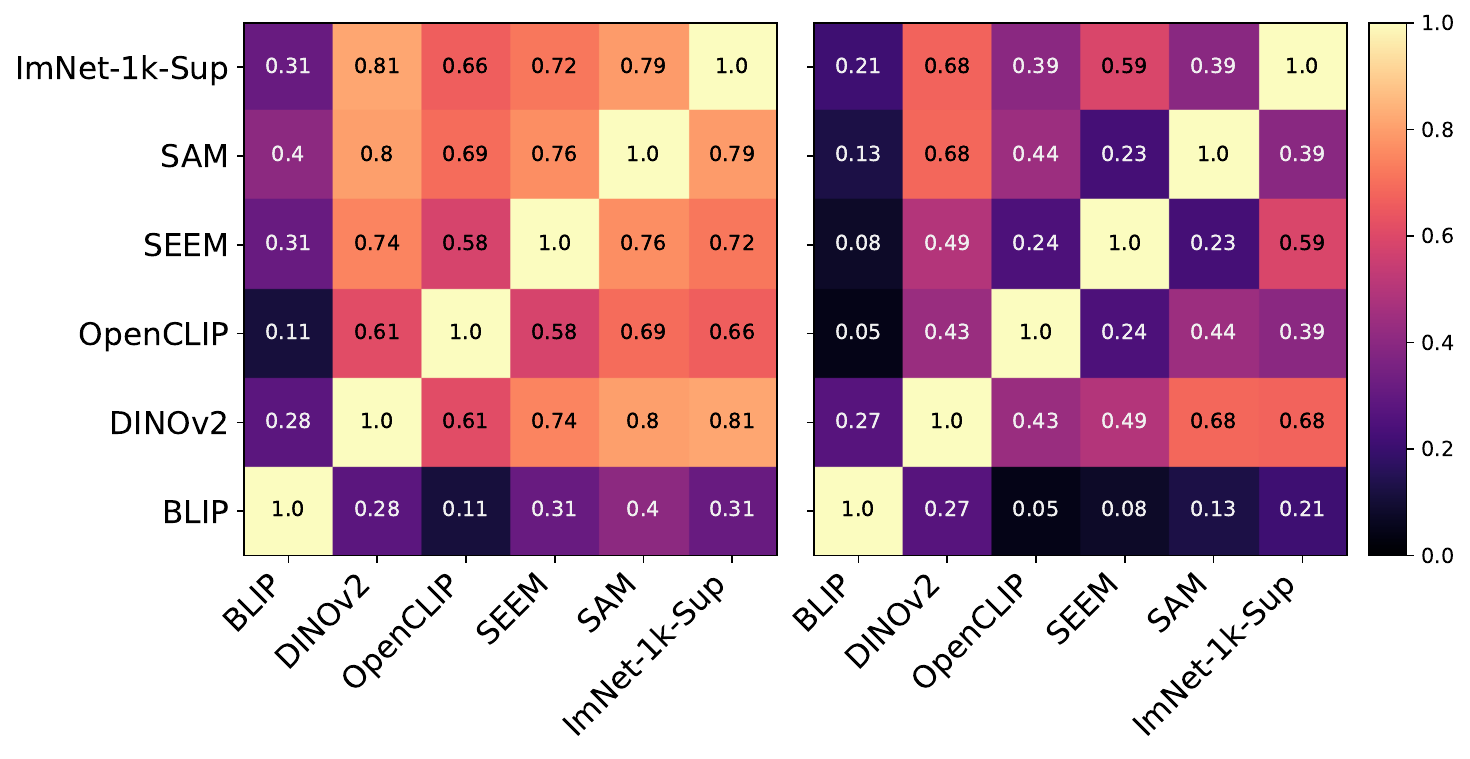} 
\end{tabular} 
\vspace{-4mm}
\caption{CKA of the last layer among the different foundation models.} 
\label{fig:beloved_fig9}
\vspace{-4mm}
\end{figure}

\noindent\textbf{Decapitating the foundation.}
Expanding on the concept that deeper layers tend to capture high-level features which may be less relevant for transfer learning, we explored what happens if we use only a part of the foundation as a feature extractor and propagated the tokens to a \deit classifier.
We decapitate the foundation model and append a DeiT-B classifier, then fine-tune. 
We consider two cases: where the foundation is either kept frozen, or fine-tuned along with the classifier.
Better performance is observed when the foundation is fine-tuned along with the classifier.
In this case, a clear trend is observed, indicating that incorporating more layers leads to improved performance (Figure \ref{fig:dropping} top). 
On the other hand, when the foundation model is used solely as a feature extractor without adaptation (Figure \ref{fig:dropping}  bottom), it appears that relying on features from the early layers yields benefits, as they tend to capture more low-level (and more general) image information. 
The optimal point varies according to the dataset and model, with block 5 performing well for \ddsm and different optimal points for \aptos depending on the model.

\noindent\textbf{Training and inference time.} 
We also evaluate the training and inference times of the different models, as shown in Figure \ref{fig:training_test}. 
Training a linear classifier on a frozen foundation took considerably longer compared to other configurations. Among the different foundation models, \dinotwo showed similar training times to the baseline \ImNet, while the others experienced increased training times, with \seem and \blip being the slowest when unfrozen.
As for the inference times, adding a full \deit classifier on the foundation results in a notable increase in inference time. However, \seem is the fastest model in terms of inference as expected due to it's progressive reduction in number of spatial tokens throughout the network. 

\noindent\textbf{Model Complexity and Image resolution.} 
Finally we analyze the top-performing model's performance by adjusting its complexity in Figure \ref{fig:sizes}.
For \isic, using larger models as feature extractors improves performance, suggesting that features are more well-adapted with increased size.
These benefits are not so clear for \aptos and \ddsm, possibly due to the smaller size and/or larger domain shift.  
Resolution experiments in Table \ref{tab:resolution} show a universal improvement in performance when high-resolution images are used, with \dinotwo benefiting the most overall.

\subsection{Internal Representation Adaptation } \label{ssec:internal_rep}

After considering the results in Section \ref{sec:usefulness}, an intriguing question arises: 
\emph{why does stacking a complete transformer-based classifier on top of the foundation yield only marginal improvements compared to a linear head?}
To address this question, we investigate how the learned representations are shared and adapted when propagated from the foundation model to the classifier network, especially considering the variations in dataset sizes and domains. 
To do this, we use the CKA similarity metric (9) as described in Section \ref{sec:evaluation}. 
This similarity analysis was performed under two distinct scenarios:

\begin{itemize}
    \vspace{-2mm}
    \item \textbf{Figure \ref{fig:fietuned_representations}} -- Between the pre-trained foundation model and the corresponding model after fine-tuning with the addition of a linear layer.
    \item  \textbf{Figure \ref{fig:beloved_fig9}} -- Between the last layer representation among different foundation models.
    \vspace{-2mm}
\end{itemize}

In Figure \ref{fig:fietuned_representations}, we observed a trend in the adaptation of foundation models that is similar to \ImNet. 
Specifically, a significant transformation was observed in a considerable number of upper layers of the fine-tuned model. 
This transformation is dataset-specific, with the \isic dataset requiring a higher degree of adaptation.
Interestingly, the low degree of change observed in the early layers suggests that foundation models possess exceptionally rich representations, requiring minimal adaptation. 
Similar analysis (Appendix Figure \ref{fig:finetuned_caps}) indicates that a high degree of adaptation is required from the stacked classifier in this unfrozen-foundation setting.   

In Figure \ref{fig:beloved_fig9}, we can observe how the features available in the last layer differ between foundation models for different datasets. Overall, \isic leads to the highest difference among models, in line with our previous observation. Interestingly, \blip remains to be the only model to learn very dissimilar features in comparison to the rest. Moreover, \blip and \clip trained for similar tasks do not exhibit any feature similarity. 

\section{Discussion}
\label{sec:discussion}

Having considered both all-purpose foundation models and segmentation foundation models, our investigation indicates that \dinotwo serves as a solid base for transfer learning to medical tasks, requiring only a standard fine-tuning for optimal performance. 
Other models fail to outperform the baselines consistently -- with the strategy of freezing the foundation model clearly failing.
Interestingly, the compact \seem model competes effectively with its larger counterparts.

We speculate on why foundation models excel in natural domain images but under perform in our tests. The architectural differences between \sam and \seem, which are both pre-trained for segmentation, might lead to varied feature hierarchies. While both \clip and \blip use text-guided pretraining, their objectives differ, potentially affecting representation quality with image-only input. Notably, \dinotwo, without extra decoders or text conditioning, outperforms. This suggests that pretraining methods and architecture significantly influence transfer learning success.

We obtained the best performance when the foundation is fine-tuned, and show that employing a linear head generally yields equal or improved performance compared to using an appended classifier. When the foundation is frozen, stacking a classifier is necessary to get the best performance, but it is often inferior to the current standard practice of \imagenet pretraining (\ImNet). Despite poor $k$-NN performance, pre-trained \cnn's match their counterparts in a supervised setting.

Within the foundation model, our analysis shows that early layers see a substantial degree of feature reuse while adaptation of the deeper layers to the target task is critical -- if the later layers are not adapted to the task they may impede performance.

\vspace{2mm}
In addition, we note the following key findings:

\begin{itemize}
\item Foundation models exhibit rich features that can be reused or adapted for medical image classification tasks, as demonstrated by the $k$-NN evaluation protocol. This is likely influenced by the large collection of natural images foundation models are trained on.
\item Freezing the foundation models often leads to a drop in performance, indicating that the high-level features generated by the foundation models are not well-suited for medical tasks.
The feature similarity analysis supports this observation, showing the emergence of novel high-level features in the last layers of unfrozen foundation models during fine-tuning.
\item Compared to the other foundation models, fine-tuning \dinotwo demonstrated a notably faster convergence to achieve high levels of performance. This phenomenon can be plausibly attributed to fewer layers requiring adaptations which can be observed from the feature similarity maps.
\item The correlation between the foundation model's training data and its transfer ability is unclear. However, a trend indicates improved performance with larger fine tuning datasets like Chexpert.
\item The role of the \emph{[cls]} token is unclear. While \dinotwo did perform best overall, the lack of a \emph{[cls]} token did not prevent \seem from performing nearly as well.
\item High resolution fine tuning leads to improved performance in both, models pretrained at high and low resolution. Additionally, the base model tends to strikes the optimal balance between performance and model complexity.  
\end{itemize}

\section{Conclusions}

In summary, the main finding of this study is that modern foundation models, notably \dinotwo, serve as a solid base for transfer learning to medical tasks with minimal fine-tuning. 
The specific pretraining scheme and architecture are crucial factors affecting performance, but already today we have foundation models that dethrone \imagenet pretraining for medical classification tasks. 
Looking to the future, as newer and larger foundation models continue to emerge, and with increasing demand for AI in medical applications, further research and exploration are needed to enhance the adaptability and effectiveness of foundation models in the medical domain. 
These advancements hold the potential to significantly improve medical image analysis and contribute to improved healthcare outcomes.

\paragraph{Acknowledgements.} This work was supported by the Wallenberg AI, Autonomous Systems and Software Program (WASP). We acknowledge the use of Berzelius computational resources provided by the Knut and Alice Wallenberg Foundation at the National Supercomputer Centre. 

{\small
\bibliographystyle{ieee_fullname}
\bibliography{References}
}
\newpage

\twocolumn[
{\center\baselineskip 16pt
    \vskip .25in{\Large\bf
    Supplementary Material for Are Natural Domain Foundation Models Useful for Medical Image Classification? \par
}\vskip .5in}
]

\begin{appendices}
\counterwithin{figure}{section}
\counterwithin{table}{section}
\section{Additional CKA analysis}

\begin{strip}
\centering
\scriptsize
\begin{tabular}{@{}c@{\hspace{2mm}}c@{}}
Frozen foundation &
Unfrozen foundation
\\
\includegraphics[width=0.5\linewidth]{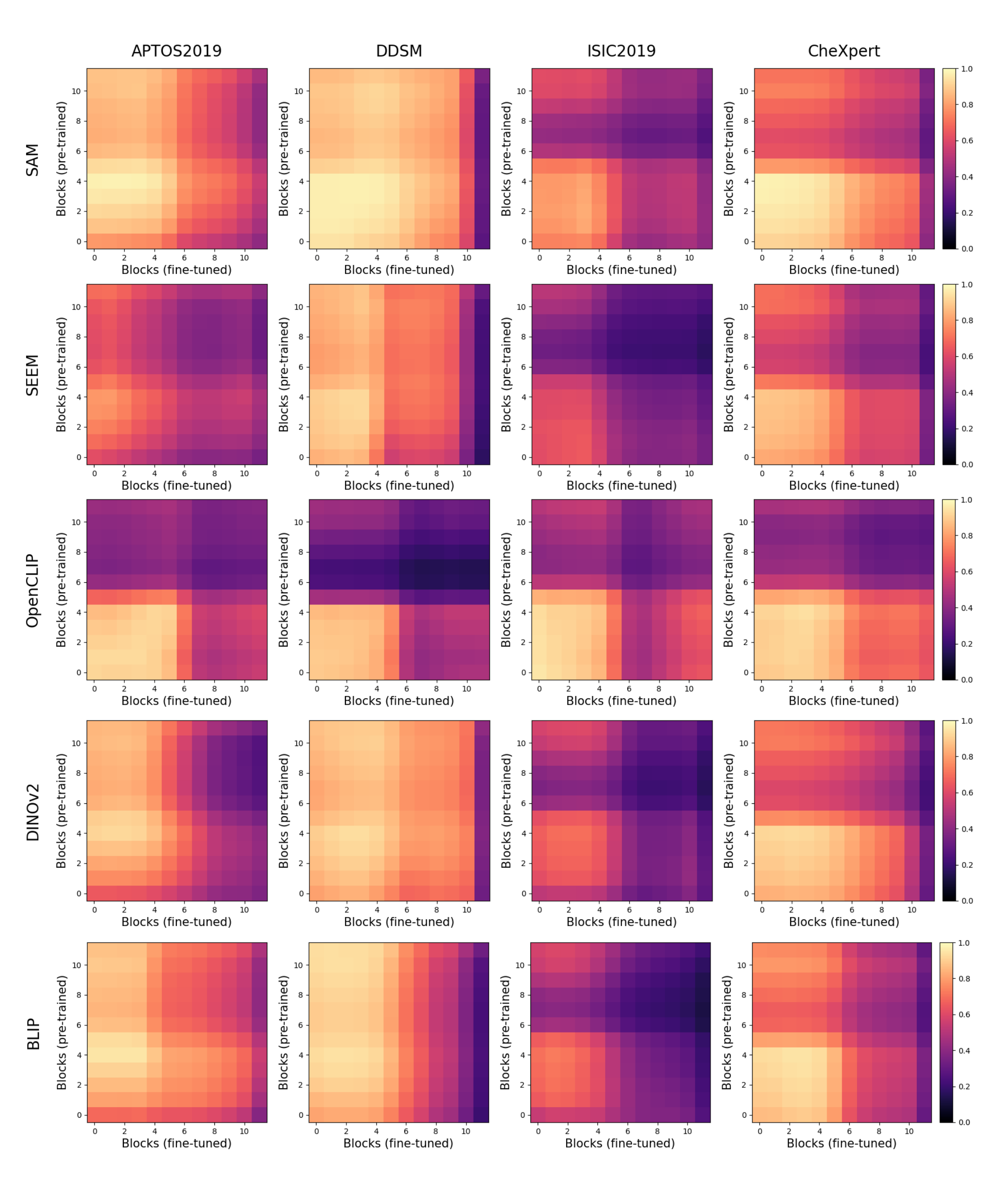} &
\includegraphics[width=0.5\linewidth]{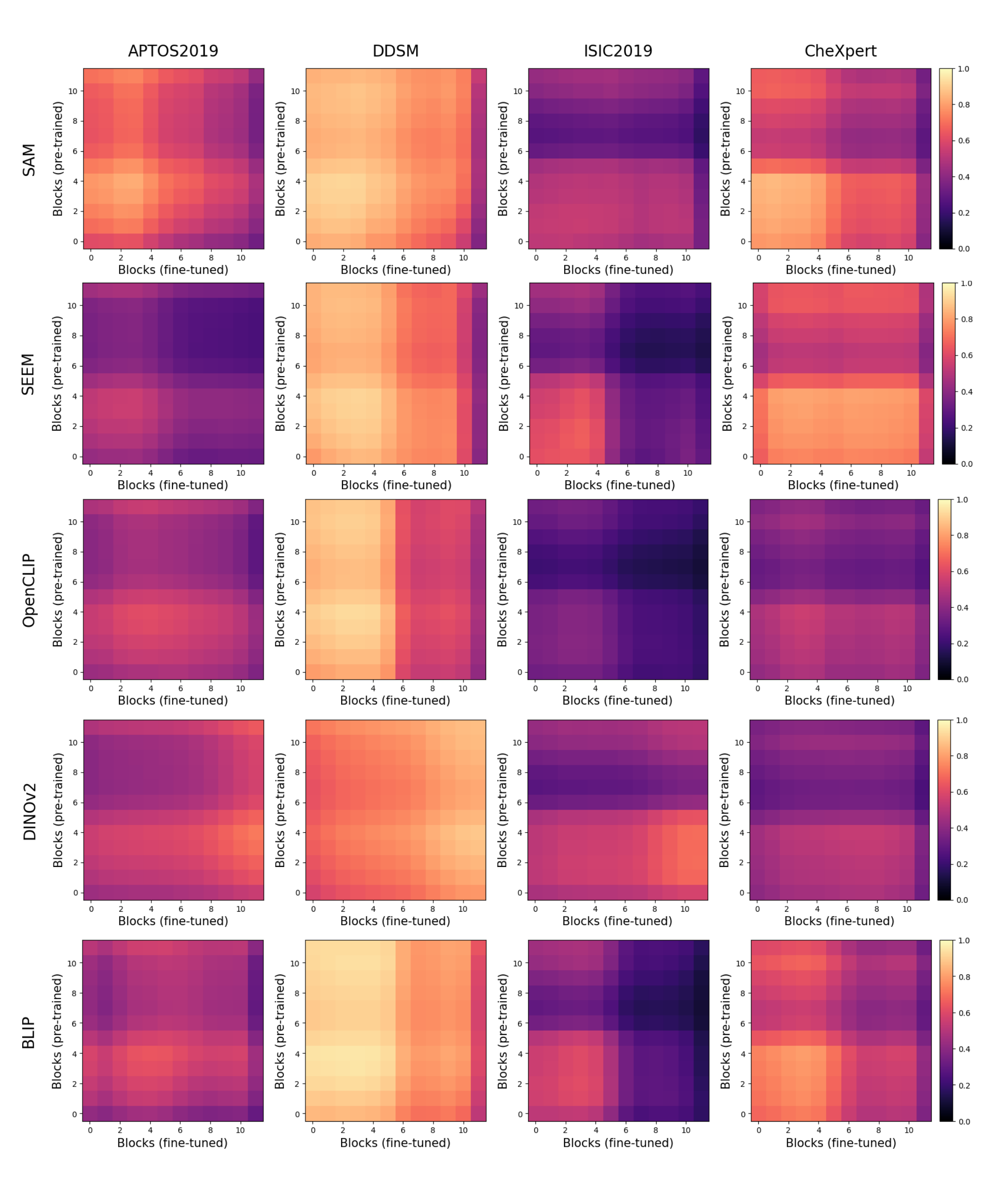} \\
\end{tabular} 
\vspace{-2mm}
\captionof{figure}{Layerwise CKA similarities between the pre-trained and fine-tuned \deit classifier appended on top of different foundation models. The heat maps are generated for two scenarios, frozen and unfrozen foundation.}
\label{fig:finetuned_caps}
\end{strip}

\clearpage
\twocolumn[
{
\section{In-domain pretraining}
\label{apx:ssl}
We investigate the impact of in-domain self-supervised pretraining. 
Specifically, we employ \dino \cite{caron2021emerging} and pretrain \resnetOneFiftyTwo models on the downstream datasets. 
We follow the training recipe described in \cite{matsoukas2021time}, followed by supervised fine-tuning as in Section \ref{sec:methods}. We report the results in Table \ref{tab:supp_frozen_table} and Table \ref{tab:supp_unfrozen_table}.
For comparison, we also present the results of the \resnetOneFiftyTwo baseline, pretrained on \imagenet-1k, and \dinotwo, the top-performer in this work.
When the foundation models are kept frozen, in-domain self-supervision significantly outperforms the other two pretraining strategies.
This is a rather expected result, given that the model has adapted its features on the downstream datasets.
Surprisingly, we find that when the full model is fine-tuned -- as typically followed in practice -- \dinotwo outperforms even its self-supervised \resnetOneFiftyTwo counterpart.

}]

\begin{strip}
\centering
\small
\vspace{-8mm}
\captionof{table}{Frozen foundation models.} 
\begin{tabular}{l@{\hskip 5pt}ccccc}
\toprule
\textbf{Foundation} & &
\textbf{APTOS2019}, $\kappa \uparrow$  &
\textbf{DDSM}, AUC $\uparrow$ &
\textbf{ISIC2019}, Rec. $\uparrow$ & 
\textbf{CheXpert}, AUC $\uparrow$
\\

(Frozen) & &
$n =$ 3,662  & 
$n =$ 10,239 &
$n =$ 25,333 & 
$n =$ 224,316 \\
\midrule
\resnetOneFiftyTwo -- pretrained on \imagenet-1k & &
0.824 $\pm$ 0.003 &
0.883 $\pm$ 0.001 &
0.461 $\pm$ 0.027 &
0.712 $\pm$ 0.000 
\\
\resnetOneFiftyTwo -- self-supervised with \dino & &
0.855 $\pm$ 0.004 &
0.920 $\pm$ 0.001 &
0.610 $\pm$ 0.020 &
0.708 $\pm$ 0.000 
\\
\dinotwo & &
0.881 $\pm$ 0.002 &
0.905 $\pm$ 0.001 &
0.569 $\pm$ 0.012 &
0.722 $\pm$ 0.000 

\\[0.5em]
\bottomrule
\end{tabular}

\label{tab:supp_frozen_table}

\centering
\captionof{table}{Unfrozen foundation models.} 
\small
\begin{tabular}{l@{\hskip 5pt}ccccc}
\toprule
\textbf{Foundation} & &
\textbf{APTOS2019}, $\kappa \uparrow$  &
\textbf{DDSM}, AUC $\uparrow$ &
\textbf{ISIC2019}, Rec. $\uparrow$ & 
\textbf{CheXpert}, AUC $\uparrow$
\\

(Unfrozen) & &
$n =$ 3,662  & 
$n =$ 10,239 &
$n =$ 25,333 & 
$n =$ 224,316 \\
\midrule
\resnetOneFiftyTwo -- pretrained on \imagenet-1k & &
0.899 $\pm$ 0.002 &
0.960 $\pm$ 0.003 &
0.817 $\pm$ 0.007 &
0.807 $\pm$ 0.000 
\\
\resnetOneFiftyTwo -- self-supervised with \dino & &
0.898 $\pm$ 0.006 &
0.954 $\pm$ 0.004 &
0.818 $\pm$ 0.002 &
0.810 $\pm$ 0.000 
\\
\dinotwo & &
0.909 $\pm$ 0.009 &
0.966 $\pm$ 0.003 &
0.859 $\pm$ 0.007 &
0.812 $\pm$ 0.001 

\\[0.5em]
\bottomrule
\end{tabular}

\label{tab:supp_unfrozen_table}
\end{strip}

\end{appendices}

\end{document}